\documentclass[conference]{IEEEtran}

\IEEEoverridecommandlockouts

\usepackage{cite}
\usepackage{amsmath,amssymb,amsfonts}
\usepackage{algorithmic}
\usepackage{graphicx}
\usepackage{textcomp}

\usepackage{times}

\usepackage{multicol}
\usepackage[bookmarks=false]{hyperref}

\everypar{\looseness=-1} 
\usepackage[table]{xcolor}
\def\BibTeX{{\rm B\kern-.05em{\sc i\kern-.025em b}\kern-.08em
    T\kern-.1667em\lower.7ex\hbox{E}\kern-.125emX}}
\usepackage{pgfplotstable}

\pgfplotstableset{
 /color cells/min/.initial=0,
  /color cells/max/.initial=1000,
  /color cells/textcolor/.initial=,
  color cells/.style={
    postproc cell content/.append code={%
          \pgfkeys{/color cells/.cd,#1}%
      \pgfkeysgetvalue{/pgfplots/table/@preprocessed cell content}\value%
      \ifx\value\empty%
      \else%
        \pgfmathfloatparsenumber{\value}\pgfmathfloattofixed{\pgfmathresult}%
        \let\value=\pgfmathresult%
        \pgfplotscolormapaccess%
            [\pgfkeysvalueof{/color cells/min}:\pgfkeysvalueof{/color cells/max}]%
            {\value}{\pgfkeysvalueof{/pgfplots/colormap name}}%
        \pgfkeysgetvalue{/pgfplots/table/@cell content}\typesetvalue%
        \pgfkeysgetvalue{/color cells/textcolor}\textcolorvalue%
        \toks0=\expandafter{\typesetvalue}%
        \edef\temp{\noexpand\pgfkeyssetvalue{/pgfplots/table/@cell content}{%
            \noexpand\cellcolor[rgb]{\pgfmathresult}%
            \noexpand\definecolor{mapped color}{rgb}{\pgfmathresult}%
            \ifx\textcolorvalue\empty\else\noexpand\color{\textcolorvalue}\fi%
            \the\toks0%
          }%
        }%
        \temp%
      \fi%
      }%
  }%
}%

\usepackage{soul}
\usepackage{url}
\usepackage[utf8]{inputenc}
\usepackage{graphicx}

\usepackage{amsmath}
\usepackage{booktabs}
\usepackage{helvet} 
\usepackage{courier}  
\usepackage{graphicx} 
\usepackage{graphicx}  
\usepackage{times}
\usepackage{latexsym}
\usepackage{url}
\usepackage{microtype}

\usepackage{times}  
\usepackage{helvet}  
\usepackage{courier}  
\usepackage{url}  
\usepackage{graphicx}  
\usepackage{amssymb}
\usepackage{amsmath}
\usepackage{textcomp}	
\usepackage{graphicx}
\usepackage{grffile}
\usepackage{ragged2e}

\usepackage{cleveref}
\crefname{section}{s}{ss}
\crefname{section}{s}{ss}
\crefname{table}{Table}{}
\crefname{figure}{Fig.}{}
\crefname{algorithm}{Alg.}{}
\crefname{equation}{Eq.}{}
\crefname{appendix}{Appendix}{}
\crefformat{section}{\S#2#1#3}  

\usepackage{booktabs}
\usepackage[vcentering,dvips,margin=1in]{geometry}
\usepackage{multirow}
\usepackage{sidecap}
\sidecaptionvpos{figure}{c}
\usepackage{adjustbox}
\usepackage{listings}
\usepackage[official]{eurosym}
\usepackage{float}
\usepackage{color}
\usepackage{soul}			
\usepackage[hang, flushmargin]{footmisc} 
\usepackage{array}			
\usepackage{balance}		
\usepackage{xspace}			
\usepackage{lipsum}
\usepackage[flushmargin]{footmisc}
\usepackage{quotes}
\usepackage{todonotes}
\usepackage{amssymb}
\usepackage{amsmath}
\usepackage{amsthm}
\usepackage{soul}
\usepackage{ulem}
\let\labelindent\relax
\usepackage[shortlabels,inline]{enumitem}

\newcolumntype{L}[1]{>{\raggedright\arraybackslash}m{#1}}
\newcolumntype{C}[1]{>{\centering\arraybackslash}m{#1}}
\newcommand{\pseudosection}[1]{\vspace{2ex}\noindent \textbf{{#1.}}}
\usepackage{enumitem}
\usepackage{wrapfig}

\usepackage{caption}
\usepackage{etoolbox}
\makeatletter
\patchcmd{\@makecaption}
  {\scshape}
  {}
  {}
  {}
\makeatletter
\patchcmd{\@makecaption}
  {\\}
  {.\ }
  {}
  {}
\makeatother
\def\tablename{Table}

\newcommand{\Token}[1]{\textit{#1}}
\newcommand{\Characteristic}[1]{\textsc{#1}}

\newcommand{\citet}[1]{\citeauthor{#1}~(\citeyear{#1})}
\newcommand{\citep}[1]{\cite{#1}}

\makeatletter
\newcommand{\linebreakand}{%
  \end{@IEEEauthorhalign}
  \hfill\mbox{}\par
  \mbox{}\hfill\begin{@IEEEauthorhalign}
}
\makeatother

\begin{document}

\title{Sampling Approach Matters: Active Learning \\for Robotic Language Acquisition}

\author{\IEEEauthorblockN{Nisha Pillai}
\IEEEauthorblockA{\textit{Univ. of Maryland, Baltimore County}}
\and
\IEEEauthorblockN{Edward Raff}
\IEEEauthorblockA{\textit{Booz Allen Hamilton}\\\textit{Univ. of Maryland, Baltimore County}}
\linebreakand 
\IEEEauthorblockN{Francis Ferraro}
\IEEEauthorblockA{\textit{Univ. of Maryland, Baltimore County}}
\and
\IEEEauthorblockN{Cynthia Matuszek}
\IEEEauthorblockA{\textit{Univ. of Maryland, Baltimore County}}
}

\maketitle







\title{\thistitle}

\begin{abstract}
%
Ordering the selection of training data using \textit{active learning} can lead to improvements in learning efficiently from smaller corpora. We present an exploration of active learning approaches applied to three grounded language problems of varying complexity in order to analyze what methods are suitable for improving data efficiency in learning. We present a method for analyzing the complexity of data in this joint problem space, and report on how characteristics of the underlying task, along with design decisions such as feature selection and classification model, drive the results. We observe that representativeness, along with diversity, is crucial in selecting data samples.
\end{abstract}

\section{Introduction}
\label{sec:intro}
In grounded language theory, the semantics of language are given by how symbols connect to the underlying real world---the so-called ``symbol grounding problem''~\cite{Harnad1990}. For example, we want a robotic system that sees an eggplant (a set of visual percepts from the real world) to ground the recognition object to a canonical symbol for `eggplant.' When a user asks "Please grab me the eggplant," the robot should ground the natural language \textit{word} "eggplant" to the same \textit{symbol} that denotes the relevant visual percepts. Once both language and vision successfully ground to the same symbol, it becomes feasible for the robot to complete the task. We learn this connection by using physical sensors in conjunction with language learning: paired language and perceptual data are used to train a joint model of how linguistic constructs apply to the perceivable world. 

Machine learning of grounded language often demands large-scale natural language annotations of things in the world, which can be expensive and impractical to obtain. It is not feasible to build a dataset that encompasses every object and possible linguistic description. Novel environments will require symbol grounding to occur in real time, based on inputs from a human interactor. Learning the meanings of language from unstructured communication with people is an attractive approach, but requires fast, accurate learning of new concepts, as people are unlikely to spend hours manually annotating even a few hundred samples, let alone the thousands or millions commonly required for machine learning.

In this work we study \textit{active learning}, in which a system deliberately seeks information that will lead to improved understanding with less data, to minimize the number of samples/human interactions required. The field of active learning typically assumes that a pool of unlabeled samples is available, and the model can request specific example(s) that it would like to obtain a label for. By having the model select the most informative data points for labeling, the number of samples that need to be labeled is reduced. This maps to the goal of human-robot learning with minimum training data provided by the human. Furthermore, active learning can be part of a pipeline with other few-shot learning methods~\cite{Gavves_2015_ICCV}. 

However, active learning is not a magic bullet. When not carefully applied, it does not outperform sequential or random sampling baselines~\cite{ramirez2017active}. Thoughtful selection of suitable approaches for problems is required. While active learning has been used for language grounding
~\cite{PillaiRSS2016workshopActive,padmakumar:emnlp18},
to the best of our knowledge, we present the \textbf{first broad exploration of the best methods for active learning for grounding vision-language pairs}. %
In this paper, our focus is on developing guidelines by which active learning methods might be appropriately selected and applied to vision-language grounding problems. We test different active learning approaches on grounded language problems of varying linguistic and sensory complexity, and use our results to drive a discussion of how to select active learning methods for different grounded language data acquisition problems in an informed way.

We consider the grounded language task of learning novel language about previously unseen object types and characteristics. Our emphasis is on \textbf{determining what methods can reduce the amount of training data} needed to achieve performance consistent with human evaluation. Primarily, we address five relevant questions concerning characteristic-based grounded language learning:
\begin{enumerate*}[(1)]
\item How much do active learning techniques help when learning with limited data?
\item Do different active learning techniques, e.g., pool-based vs. uncertainty-based approaches, lead to noticeable differences in performance?
\item Are the methods robust across both neural and non-neural features and classifiers?
\item How important are the characteristics of the dataset? and 
\item How much does incorporating some seed language affect the performance?
\end{enumerate*}
We make conclusions with respect to these questions in \cref{sec:results}. %
In addition to addressing the above research questions, we verify how generalizable these learning techniques are beyond characteristic-based grounding.  

We find that a right ordering of training data makes it possible to learn successfully from significantly fewer descriptions in most cases, but also that the active learning methodology chosen is specific to the nature of the learning problem. Our main contribution is a \textbf{principled analysis of using active learning methods as unsupervised data sampling techniques} in language grounding with a discussion of what aspects of those problems are relevant to approach selection. While our contributions are primarily analytic rather than algorithmic, we argue they address a critical need within grounded language understanding, an active research area in which questions of efficiency and data collection are widespread, and have the potential to support additional algorithmic developments.

\section{Related Work}\label{sec:related}

Grounded language learning has been successful in learning to follow directions~\cite{artzi13,anderson2018vision}, generating referring expressions~\cite{Shridhar-RSS-18}, visual storytelling~\cite{huang-2016-storytelling}, video grounding~\cite{Shi_2019_CVPR} and understanding commands~\cite{chai2018language}, among others. Parsing can be grounded in a robot's world and action models, taking into account perceptual and grounding uncertainty~\cite{TellexAAAI2011,walter2014framework,MatuszekIJCAI2018,thomason1}. 
or language ambiguity~\cite{ChenAAAI2011,chaplot2018gated}. 
The problem space considered in this paper assumes that there are no pre-existing models of language or objects in the world---an agent is learning from novel language about previously unseen objects~\cite{MatuszekICML2012,tucker2017learning}, making the evaluation more broadly applicable. 

Active learning has been applied successfully to a number of problems~\cite{10.1007/978-3-319-02675-6_46, short2019sail, bullard2019active}, providing performance improvements in areas as diverse as learning from demonstration~\cite{cakmak2010designing,bullard2018towards}, following directions~\cite{hemachandra2015information}, and learning about object characteristics~\cite{thomason2017opportunistic}. A well-chosen active learning approach can reduce the number of labels required for grounded language learning~\cite{PillaiRSS2016workshopActive,AAAI1817265}, but raises questions of what queries to ask and when to ask them~\cite{cakmak12,TellexThakerDeitsEtAl2013,AAAI1714452}. 

Advances in active learning techniques have improved the ability to find the most useful data points. Unsupervised learning techniques, such as subspace clustering, have been shown to find influential points from a cluster~\cite{peng2019subspace}. A hybrid method that connects active learning and data programming~\cite{nashaat2018hybridization} has shown improvements in the reduction of noisy data in large scale workspaces~\cite{chen2020autodal}. Similar to our work, active learning approaches~\cite{Gudovskiy_2020_CVPR,NIPS2019_8457, NIPS2019_8500} have been effective while training biased and highly varied datasets. Similarly, researchers have put effort into utilizing different active learning methods depends on the complexity of the problem~\cite{NIPS2019_8831}. Also, traditional active learning methods have helped to improve performances in other tasks, such as data fault or fake news detection~\cite{homayouni2019interactive, bhattacharjee2017active}. Though we consider efficiency over time complexity, researchers have studied methods that are time efficient, especially in large scale applications~\cite{huang2017big}. Similar to our research,~\citep{chhatwal2017empirical} also compares two traditional active learning algorithms for selecting important points from a pool of training data. But we also consider distinct machine learning approaches with small scale and large scale datasets in our comparisons. Various Bayesian techniques have been used in selecting diverse points as the most influential~\cite{NIPS2019_8865} is widely popular, and we use different variants of  DPP to select distinct data points as our active learning technique in batch sample selection.

In this work, our goal is to perform a principled exploration of selecting what data to query for labeling~\cite{liu2019active}, using informativeness and uncertainty metrics~\cite{wang2019bounding,bullard2019active} in grounded language problems of varying complexity. We draw on existing techniques, particularly pool-based learning~\cite{zhang2014beyond,kontorovich2016active,tang2019self}, uncertainty sampling
~\cite{lewis1994sequential,kong2019active}, and probabilistic sample selection~\cite{sarawagi2002interactive}. 
We take advantage of that body of research to select our set of experimental approaches, which include sample selection via Gaussian mixture models~\cite{cohn1996active,khansari2011learning} and Determinantal Point Processes (DPPs)~\cite{kulesza2012determinantal}, which have proven effective in modeling diversity~\cite{NIPS2018_7923,Wu_2018_CVPR}. Using supervised learners as the active learning techniques~\cite{DBLP:journals/corr/abs-1801-07875,tang2019self} are not suitable for our current study since we concentrate on building a language model without prior knowledge~\cite{krawczyk2019adaptive}. 

Our work is most closely related to that of Thomason et al.~\citep{thomason2018guiding}, who incorporate `opportunistic' active learning in a system that learns language in an unstructured environment~\cite{thomason2017opportunistic,padmakumar:emnlp18}. However, that work focuses on opportunistically querying for labels whenever annotators are present; this work, in contrast, is focused on exploring the best way of selecting good choices from a large range of possible queries, reflecting the assumption that opportunities to query users will often be severely limited.


\begin{figure}[b!]
\centering
\begin{tabular}{>{\raggedright\arraybackslash}m{.11\columnwidth}|m{.15\columnwidth}|>{\raggedright\arraybackslash}m{.54\columnwidth}}
\small \textbf{Type} & \small \textbf{Image} & \small \textbf{English annotation} \\ \hline\hline
\small color & \includegraphics[width=0.14\columnwidth]{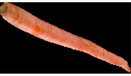} & \small This is an \textbf{orange} object. \\ \hline
\small shape & \includegraphics[width=0.14\columnwidth]{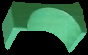} & \small This looks like a green \textbf{upside} \textbf{down} \textbf{C} \textbf{shape}. \\ \hline
 \small {\shortstack{object\\ type}} & \includegraphics[width=0.14\columnwidth]{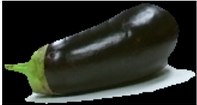} & \small This is an \textbf{Italian Eggplant}. It is firm and dark purple when ripe. 
 \end{tabular}
\caption{RGB-D sensor data and descriptions~\protect\cite{PillaiAAAI2018}. Each concept was used by multiple annotators to describe each of the corresponding images, showing the noise and variability of human descriptions. 
}
\vspace{-1ex}
\label{fig:token_images}
\end{figure}

\section{Approach}\label{sec:approach}

For different active learning methods, we learn associations between RGB-D images (color+Depth) of objects in a dataset and the language that describes them. 
The task is then to find concepts that have a grounded meaning, create lexical terms in an underlying formal meaning representation, and learn visual classifiers that correctly identify things that are referred to in later language interpretation tasks. 

At a high level, we ground language by learning characteristic-specific classifiers such as color, shape, and object for a concept. Consistent with previous work~\cite{PillaiAAAI2018}, the different types of concepts are obtained from human-provided descriptions of selected objects. In this approach, each concept is associated with a (learned) classifier, and all selected objects described by that concept are used as training data for that classifier. We rely on existing datasets and classification approaches for the actual grounding. We note that the evaluations done in this work are intended to \textit{compare} the success of different active learning approaches for the same problem. 

We limit training data to a single description of each object to mimic the limited training available from human interactions. In order to perform replicable experiments, we use active learning approaches in which objects (and associated training and evaluation information, such as descriptions and identified concepts) are drawn from a pre-existing pool of data, rather than obtained \textit{de novo} through human interaction. In our primary experiments we vary the active learning approach used to select new descriptions of objects to add to the training pool. We additionally experiment with different features, and classification techniques. Because our problem focuses on choosing objects to obtain labels for, this is consistent with the task of asking a person for a description of a particular object, but allows us to perform larger-scale and more replicable experiments.

Our goal is to explore the data selection decisions in limited settings to improve performance at the early stages. It is not to improve absolute learning performance; using a novel or complex approach runs the risk of introducing poorly understood confounding factors.

\subsection{Data: Corpora, Concepts, and Features}\label{subsec:data}

We use two existing datasets for learning from descriptions: the UMBC dataset~\citep{pillaiidentifying2017}, which contains 72 objects (see \cref{fig:token_images}), and the UW RGBD+ dataset~\citep{lai2011dataset}, which contains 300 objects. Each object instance has multiple associated language descriptions. We follow existing literature~\cite{richards2019learning,Kery2019ROMAN, kery2019esta} on learning to understand language referring to different types of characteristics: \Characteristic{color}, \Characteristic{shape}, and \Characteristic{object type}. The corpora consist of Kinect2 depth images of objects, paired with human descriptions. The UMBC object dataset contains 8 color, 9 shape, and  18 object characteristics, while the UW RGBD+ dataset includes 14 color, 13 shape, and 51 object characteristics. Shape concepts are reported only in 9.5\% descriptions of all  300 object UW RGBD+ dataset annotations. In the UMBC dataset, 53\%, 14\%, and 73\% of annotations reported color, shape, and object concepts.
%

\textit{\textbf{Language.}}~~The UMBC dataset contains approximately 430 natural language object descriptions, while the UW RGBD+ dataset contains 1500 descriptions; all were obtained via Amazon Mechanical Turk. For each image of an object, a single description is randomly selected to pair with an image it describes; this is intended to replicate the limited labels available from human interaction. We opt for the simplicity of learning meanings for individual concepts, based on past effectiveness of this approach~\cite{mei2016listen,PillaiAAAI2018}. Following this previous work, we first convert descriptions into language \textit{concepts}, removing common stop words and lemmatizing the remainder. We then identify meaningful, relevant, and representative concepts by applying tf-idf~\cite{salton1986introduction}, which yields concepts such as `banana' and `yellow', while rejecting those such as `object' and `look' (See section \ref{sec:results} for details).

\textit{\textbf{Sensor Data.}}~~Physical context for language grounding is provided by depth and color images of each object, taken with an RGB-D camera mounted on a robot platform (\cref{fig:token_images}). From each RGB-D image, we extract perceptual features $\eta_{\Characteristic{trait}}$ for each different type of characteristic. We use two different kinds of visual features for learning. In the first, a kernel descriptor-based approach, we use the average RGB values for color, HMP-extracted kernel descriptors~\cite{lai2011dataset} for shape, and a combination of the two for objects. In our second approach, we use a convolutional neural network, Neural Architecture Search Network (NASNetLarge)~\cite{zoph2018learning}, with pretrained ImageNet~\cite{deng09} weights for extracting a 1024 dimension feature vector.

Across all combinations we found that \Characteristic{color} is relatively easy to learn; \Characteristic{shape}, which depends in part on camera angle and is less likely to be mentioned, is more difficult; and \Characteristic{object type} is the finest grained, with the highest visual complexity. 

\subsection{Learning Concept Classifiers}\label{subsec:learning_groundings}

The task we use to test approaches is to learn associations between perceptual inputs and descriptive concepts. Once perceptual features are extracted from the images, a visual classifier for each characteristic is learned. These classifiers are trained using every image that has been described with a concept \textit{and} selected by an active learning method. %



Given an instance $x_i$ and a characteristic-specific perceptual representation $\eta_{\Characteristic{trait}}(x_i)$, we learn characteristic-specific probabilistic binary classifiers for each concept,
$
p_{\Characteristic{trait}}(w_{\Token{concept}}\ |\ \eta_{\Characteristic{trait}}(x_i)),
$
where $w_{\Token{concept}} \in \{0, 1\}$ represents the probability of $x_i$'s characteristic \Characteristic{trait} being described as \Token{concept}. %
Note that this problem is two-fold: we must learn how to both describe objects properly, and how to \textit{avoid} characterizing objects in a way that does not make sense. %
We use logistic regression (LR) as our primary classifier type $p_{\Characteristic{trait}}$ (see \cref{subsec:classifier} for the impact of this decision) and extract characteristic-specific features $\eta_{\Characteristic{trait}}$. %

\subsection{Core Sampling Methods}\label{subsec:initial}

Intuitively, we want our algorithms to select preferentially the most informative and diverse objects for labeling from the pool of unlabeled objects. Driven by both long-standing and recent findings in active learning~\cite{cohn1996active,NIPS2018_7923,settles2012active,wang2019comparison}, we use probabilistic clustering---and point process modeling in particular---as active learning strategies. Because our data is inherently noisy, we found in our early experiments that variations on Gaussian mixture models (GMMs) and determinantal point processes (DPPs) were robust selection algorithms. GMMs accommodate mixed membership, and soft cluster assignments allow us to model uncertainty. We select parametric methods in our learning techniques as they are statistically stable~\cite{8594676} compared to nonparametric models. We therefore focus on GMM- and DPP-based approaches, applied to visually grounded object features, in order to select the most informative points from a set of unlabeled instances. 

As we focus on learning from limited data, we do not consider deep learning approaches, which generally operate best over large datasets. Across all of our experiments, we examine five different active learning models: four pool-based methods (GMM Max Log Density Based, 
VL-GMM,\footnote{VL-GMM is included to show the difference between vision-only \textit{vs.} vision-language clustering-based learning, and so does not occur in other reported results.} and DPP), and one uncertainty-based (GMM Log Density) method. %
We introduce a structured DPP (GMM-DPP)-based active learning technique, a novel approach for the grounded language problem. We compare these variants of active learning strategies with a random sampling baseline across our three characteristics (color, shape, and object). Although initial experiments considered entropy-based sampling methods (computed by our GMM's posterior entropy), these were found to perform substantially worse than those listed, and subsequent experiments did not include them. For all GMM approaches, we select the number of components $C$ empirically using four-fold cross-validation. In GMM based methods, we compared the test performance with the number of components ranging from 5 to 35 and received the best results with 15 components. 
In GMM-based pool sampling experiments, we cluster instances using their informativeness and rank the instances according to their learned conditional densities.

Our methods select instances which are informative and diverse by querying from all $N$ items at once. This is also called querying in ``batch mode'' and has been applied successfully in the past~\cite{sarawagi2002interactive,chattopadhyay2013batch}. We draw from an existing pool of human-provided descriptions, rather than explicitly seeking new labels via interaction, to enable broader and more repeatable experiments.


\textit{Max Log-Density-Based GMM Sampling:} This model uses a $C$-component GMM to cluster unique image features and rank them according to their maximum multivariate densities from the unlabeled data pool. Those with greater density are selected as they are potentially more informative. We used 15 Gaussian components (selected empirically as a hyperparameter), initialized the mixing weights and Gaussian parameters using $k$-means, and fit the GMM with the standard expectation maximization algorithm to learn the parameters.

\textit{DPP Sampling:}\label{kdpp}
DPPs have proven effective in modeling diversity~\cite{gong2014diverse}. We  use DPPs as a technique to find the most representative and diverse data points from the pool of data instances. This method uses the pool of all unlabeled image samples to find the most diverse data points by using a radial basis function (RBF) kernel with carefully selected parameters. %
In our setting, DPPs define a discrete probability distribution of all subsets of image data samples. If $\mathbf{X}$ is the random variable of selecting a subset of images $X$ from a larger set $\mathcal{X}$, then 
$P(\mathbf{X}=X) = \mathrm{det}(K^{(0)}_X)/\mathrm{det}(K^{(0)}_{\mathcal{X}} + I),$
where $I$ represents the identity matrix. %
Applied to all pairwise elements of $X$, the \textit{kernel} $K^{(0)}_X$ is a positive semi-definite matrix, where the $(i,j)$ element of the matrix is the value of the kernel applied to items $x_i$ and $x_j$. We use the RBF kernel, $K^{(0)}(x_i, x_j) = \exp{\left(-h \|x_i-x_j\|_2^2\right)}$, 
by cross-validating with $h \in \{100, 25, 4\}$.

\textit{GMM-DPP:} 
We combine the DPP kernel with the GMM marginal probability derived from the image samples to rank input samples based on diversity. Following Kulesza and Taskar~\citep{kulesza2010NIPS} and Affandi et al.~\citep{affandi2014learning}, we combine a DPP Kernel $K^{(0)}(x_i, x_j)$ defined on images $x_i$ and $x_j$ with individual ``quality'' scores for each of the images. %
We use $P_{\mathrm{GMM}}(x)$---the marginal probability of image $x$ according to the GMM---as the quality scores, and define a new kernel as:
\begin{equation*}
\label{eq:sdpp}
K^{(1)}(x_i, x_j)=P_{\mathrm{GMM}}(x_i)K^{(0)}(x_i,x_j)P_{\mathrm{GMM}}(x_j) %
\end{equation*}

\noindent The marginal probability modulates the diversity of the data. It allows a separate model, with its own assumptions, to help designate what data is and is not diverse. To the best of our knowledge, this is a novel kernel for grounded language learning. Similar to the GMM-based sampling approach, we used 15 Gaussian components in the GMMs and we initialized the mixing weights and Gaussian parameters using k-means.

\section{Experimental Setup}\label{sec:results}

We estimate the quality of grounded language acquisition by the predictive power of learned concept classifiers against the test objects. In \cref{tab:auc} we calculate area-under-the-curve (AUC) from the $F_1$-score performance of concept classifiers. Our baseline randomly picks images to train visual classifiers while the active learning approaches sample data points as described above. This is meant to mimic the performance of a robot asking random questions about objects in the environment.

The baseline and our active learning methods all only observe concept words from a single text description for each image. Images which are described by these words are selected as positive instances. Similarity metrics are used to find negative examples for these words~\cite{pillai2018optimal}. 
All results are averaged over 4--12 runs for each of object, shape, and color. We selected hyperparameters, such as the number of components of our GMM model empirically via cross-validation. We also selected the query size for each experiment empirically. 

\section{Results and Per-Characteristic Analysis}

The overall performance of each approach on the language learning task is shown in \cref{tab:auc}, divided into the three characteristic learning problems addressed, namely color, shape, and object. 
\subsection{In-Depth Analysis of Active Learning Performances}
The effect of active learning techniques in grounded characteristics learning is measured by comparing thre three pool-based active learning techniques described previously, with the random sampling baseline for color, shape, and object characteristics (\cref{tab:auc}). Below we will give an analysis of the results with respect to Color, Shape, and Object grounding. 

 \renewcommand{\tabcolsep}{1pt}
\begin{table}[t]
\centering
\pgfplotstabletypeset[col sep=comma,
/color cells/max=1.0,
/color cells/min=0.0,
/color cells/textcolor=black,
columns/Color/.style={color cells},
columns/Shape/.style={color cells},
columns/Object/.style={color cells},
columns/Sampling/.style={string type,column type = {p{5cm}}},
/pgfplots/colormap={whitegray}{rgb255(0cm)=(255,255,255);rgb255(1cm)=(192,192,192)},
column type={>{\fontseries{bx}\selectfont\centering\arraybackslash}c},
every head row/.style={after row=\midrule},
every column/.style={
                column type/.add={}{|}
            },
]{
Sampling,Color,Shape,Object
\textit{Baseline: Random} , 0.75 , 0.19 , 0.49
\textit{Max-Log-Density-Based GMM Pool}, 0.82 , 0.25 , 0.62
\textit{DPP Sampling} , 0.80 , 0.22 , 0.59 
\textit{{GMM - DPP Sampling}}, 0.78 , 0.27 , 0.58
}
\caption{AUC summaries for each method's $F_1$ performance, grouped by the characteristic learned. All AL techniques performed better in characteristic grounding by selecting significant points from the pool. }
\label{tab:auc}
\end{table}


\pseudosection{Color} \Characteristic{Color} is the simplest of the three categories of characteristics learned. This observation is, in part, a result of the dataset, in which objects are primarily all of one color; it is also a simpler vision problem overall. Similarly, there is little variation in the color descriptions. Most annotators used simple color names (e.g., ``red'') rather than the full range of available English terms (e.g., ``crimson''). Noisy annotations such as a carrot being described as ``purple'' and ``rose'' make the learning problem difficult. 
To train our color classifiers, we extract RGB features of the segmented object; these define $\eta_{\Characteristic{color}}$ and were shared across all approaches.

All active learning techniques outperformed a random baseline in learning groundings for color concepts. 
When neither visual percepts nor descriptive language varies widely, the primary consideration is to choose representative data quickly. DPP-based sampling methods, which by design select diverse points, also learned effective classifiers with limited data.  


\pseudosection{Shape} The second category of results, \Characteristic{shape}, is the most visually complex, and of extreme linguistic difficulty due to limited annotations. Learning shape classifiers is a comparatively complex problem, as the shape of an object varies with viewing angle. A wider variety of words is used to describe shapes but unlike describing colors, users tend not to explicitly specify objects' shapes, e.g., when asked to describe a lemon, most people say yellow, but relatively few say ``round''. %
To train shape classifiers, we extract kernel descriptors of the segmented object;
these define $\eta_{\Characteristic{shape}}$ and were shared across all approaches. 

The random sampling baseline is affected by the lack of shape tokens in the description, requiring nearly 30 descriptions to learn the first few shape words. GMM-based DPP showed a noticeable improvement in speed of learning, and also, on inspection, found distinct shape words faster than random sampling. All active learning approaches that found diverse points at earlier stages also outperformed the random baseline. 

\pseudosection{Object Type}
The next challenging grounding task considered in this work is \Characteristic{Object}---learning language that describes membership in an object class, i.e., object recognition. To train object classifiers, we extract both RGB and kernel descriptors \citep{bo2010kernel}; these define $\eta_{\Characteristic{object}}$, meaning that object recognition is treated approximately as a superset of color and shape learning.

Performance of Max-Log-Density-Based GMM Pool sampling approach is significantly better than the random baseline. We believe we observed this result because the number of classes is larger (and membership is therefore sparser) than for color and shape characteristics, reflecting the complexity of `real world' sensor data. This sparsity makes careful selection of samples particularly critical. 

\subsection{Pool Vs. Uncertainty-Based Active Learning Methods}

Uncertainty sampling methods use learned probability models to measure the uncertainty in unlabeled data points. 
\textit{Log-Density-Based GMM Uncertainty Sampling:} uses a learned GMM to pick  outliers. We select these by finding the images that have the \textit{lowest} log-density from any GMM component. We aim to select the most uncertain data points in order to get a diverse dataset.

\renewcommand{\tabcolsep}{1pt}
\begin{table}[t]
\centering
\pgfplotstabletypeset[col sep=comma,
/color cells/max=1.0,
/color cells/min=0.0,
/color cells/textcolor=black,
columns/Color/.style={color cells},
columns/Shape/.style={color cells},
columns/Object/.style={color cells},
columns/Sampling/.style={string type,column type = {p{5cm}}},
/pgfplots/colormap={whitegray}{rgb255(0cm)=(255,255,255);rgb255(1cm)=(192,192,192)},
column type={>{\fontseries{bx}\selectfont\centering\arraybackslash}c},
every head row/.style={after row=\midrule},
every column/.style={
                column type/.add={}{|}
            },
]{
Sampling,Color, Shape , Object
\textit{Baseline: Random} , 0.75 , 0.19 , 0.49
\textit{Max-Log-Density-Based GMM Pool}, 0.82 , 0.25 , 0.62
\textit{Log Density Based GMM Uncertainty}, 0.83 , 0.23 , 0.44
}
\caption{AUC summaries of $F_1$ performance for Pool and Uncertainty sampling performance, grouped by the characteristic learned. Uncertainty sampling (which depends on the feature variability) does not perform well in object grounding, which has a noisy, highly varied data pool.}
\label{tab:gmmauc}
\end{table}

Max log-density-based GMM pool based sampling (\cref{tab:gmmauc}) chooses representative data points from the unlabeled pool of objects, whereas uncertainty sampling selects the diverse points by considering outliers as the useful points. The selection depends on the variability of the features. For learning color and shape concepts, both pool- and uncertainty-based sampling performed better than the baseline. But while learning object types, the uncertainty sampling could not get required concepts from the most varied visual set and limited annotation dataset. 

We hypothesize that the deterioration of uncertainty pooling on the Object task relates to the nature of information's utility in an active learning context. As more information and descriptors become available in the Object scenario, it becomes easier for outliers to occur: points with unusual shapes and color combinations that are not well described will increase model uncertainty. Obtaining a label for an outlier may have limited utility for future data due to the precise nature of being an outlier: its behavior is inconsistent with the rest of the data. This may make uncertainty based approaches less attractive as more complex grounded language datasets become available, or indicate a need in refinement to uncertainty based approaches. 

\subsection{The Impact of Visual Features}
Convolutional Neural Network (CNN) features have been shown to be effective in learning characteristic types~\cite{xu2019explicit}. In this section, we examine how robust our active learning methods are across both neural and non-neural features. In contrast to the ``kernel descriptors'' (the RGB and HMP features used in the previous section), we extracted 1024 dimension features from the Neural Architecture Search Network (NASNetLarge), which is pre-trained on ImageNet. We refer to the NASNetLarge features as the ``CNN'' features. %

 \renewcommand{\tabcolsep}{1pt}
\begin{table}[t]
\centering
\pgfplotstabletypeset[col sep=comma,
/color cells/max=1.0,
/color cells/min=0.0,
/color cells/textcolor=black,
columns/CNN/.style={color cells},
columns/KernelDesc/.style={color cells},
columns/Sampling/.style={string type,column type = {p{5.5cm}}},
/pgfplots/colormap={whitegray}{rgb255(0cm)=(255,255,255);rgb255(1cm)=(192,192,192)},
column type={>{\fontseries{bx}\selectfont\centering\arraybackslash}c},
every head row/.style={after row=\midrule},
every column/.style={
                column type/.add={}{|}
            },
]{
Sampling,CNN,KernelDesc
\textit{Baseline: Random} , 0.53 , 0.49 
\textit{Max-Log-Density-Based GMM Pool}, 0.66 , 0.62
\textit{DPP Sampling} , 0.55 , 0.59
\textit{GMM - DPP Sampling}, 0.49 , 0.58
}
\caption{AUC summary results for each visual feature's $F_1$ performance for ``object'' characteristics. DPP, and GMM pool are able to consistently outperform the baseline with both types of visual features (non-neural kernel descriptors and CNN features).}
\label{tab:featuresauc}
\end{table}

\Cref{tab:featuresauc} shows that, similar to grounded learning with kernel descriptors, most of the active learning techniques are able to outperform the random baseline on CNN features. DPP and Max log-density-based GMM pool active learning techniques are able to pick diverse and representative points at earlier stages. 
The characteristic learning example above shows that active learning is effective in selecting meaningful and diverse points faster irrespective of the underlying visual features. %
These results also show that in a low-data regime, using a CNN over Kernel descriptors without considering the specific method of active learning used can lead to inferior results. Using CNN features with both DPP sampling approaches yields lower AUC than Kernel Descriptors. While the Max-Log-Density approach dominates in this setting, it shows why the study of the impact of features in combination with active learning is necessary. 

\subsection{Analysis with Different Classifiers}
\label{subsec:classifier}
In this section, we revisit our choice to use a logistic regression classifier for $p_{\Characteristic{trait}}(w_{\Token{concept}}\ |\ \eta_{\Characteristic{trait}}(x_i))$, and we examine how robust our active learning methods are across different classifiers. %
We consider a support vector machine (SVM) and a multilayer perception (MLP). The SVM is a well-known linear model that finds the maximum-margin hyperplane, which distinctly classifies the data samples. An MLP is a feed-forward artificial neural network that uses nonlinear activation functions. Both have been widely used for classification.


 \renewcommand{\tabcolsep}{1pt}
\begin{table}[!ht]
\centering
\pgfplotstabletypeset[col sep=comma,
/color cells/max=1.0,
/color cells/min=0.0,
/color cells/textcolor=black,
columns/LR/.style={color cells},
columns/SVM/.style={color cells},
columns/MLP/.style={color cells},
columns/Sampling/.style={string type,column type = {p{5.5cm}}},
/pgfplots/colormap={whitegray}{rgb255(0cm)=(255,255,255);rgb255(1cm)=(192,192,192)},
column type={>{\fontseries{bx}\selectfont\centering\arraybackslash}c},
every head row/.style={after row=\midrule},
every column/.style={
                column type/.add={}{|}
            },
]{
Sampling,LR,SVM,MLP
\textit{Baseline: Random} , 0.75 , 0.72 , 0.62
\textit{Max-Log-Density-Based GMM Pool}, 0.82 ,0.66 , 0.54
\textit{DPP Sampling} , 0.80 , 0.66 , 0.60
\textit{{GMM - DPP Sampling}}, 0.78 , 0.63 , 0.50
}
\caption{AUC summary results for each classifier's $F_1$ performance for ``color'' characteristics. Logistic regression is effectively able to classify the types with diverse and meaningful points.}
\label{tab:classifierauc}
\end{table}

In this experiment, we examined the ``color'' characteristic learned with the three classifiers (LR, SVM, and MLP).  In \cref{tab:classifierauc} we see that, across active learning methods, logistic regression classifiers are able to classify better than the random sampling baseline. In contrast, neither the SVM nor the MLP resulted in effective classification models when paired with active learning approaches. 
These results suggest that complex classification methods may not yield improved performance, and show the need to consider the selection of active sampling methods and downstream classifiers jointly. 

\subsection{Analysis with Different Datasets} %
In this section we examine if our techniques are effective for a large dataset that is visually and linguistically noisy and diverse. In addition to the limited features dataset, we tested our active learning techniques over a 300 object UW RGDB+ multi-colored dataset (\cref{tab:datasetauc}) for just ``color'' characteristics due to space constraints. It contained 51 objects and 1500 annotations (Sec. \ref{subsec:data}). In the UW RGBD+ dataset, not every description contains color information.  Additionally, the words used to describe the color concepts are inconsistent. Since the dataset contains fewer monochromatic objects, the visual variation is also high, making the vision-language grounding a challenging task. Even in these experiments, most of the learning techniques which selected diverse and representative points were able to perform better than a random baseline. DPP fails to rank in order of importance when the linguistic and visual data is inconsistent. These results indicate that our active learning techniques are generalizable and equally beneficial to datasets on different scales.


\renewcommand{\tabcolsep}{1pt}
\begin{table}[!ht]
\centering
\pgfplotstabletypeset[col sep=comma,
/color cells/max=1.0,
/color cells/min=0.0,
/color cells/textcolor=black,
columns/UMBC/.style={color cells},
columns/UW RGBD+/.style={color cells},
columns/Sampling/.style={string type,column type = {p{5.5cm}}},
/pgfplots/colormap={whitegray}{rgb255(0cm)=(255,255,255);rgb255(1cm)=(192,192,192)},
column type={>{\fontseries{bx}\selectfont\centering\arraybackslash}c},
every head row/.style={after row=\midrule},
every column/.style={
                column type/.add={}{|}
            },
]{
Sampling,UMBC,UW RGBD+
\textit{Baseline: Random} , 0.75 , 0.53
\textit{Max-Log-Density-Based GMM Pool}, 0.82 , 0.58
\textit{DPP Sampling} , 0.80 , 0.51 
\textit{{GMM - DPP Sampling}}, 0.78 , 0.64
}
\caption{AUC summary results for each dataset's $F_1$ performance for \Characteristic{color}. GMM pool and GMM-DPP are able to consistently outperform baseline even with a multi-colored UW RGBD+ dataset.}
\label{tab:datasetauc}
\end{table}

\subsection{The Impact of Seed Language}

So far, our methods have selected images without considering the concepts of the objects represented; in this section, we revisit that restriction and examine whether active learning methods can benefit from considering both the image and language description together. %
To do this we define a joint vision-language pool-based model that uses a combination of language informativeness and visual features to choose sample points from the data pool. We call this method \textit{VL-GMM Sampling}. We use paragraph vectors~\cite{le2014distributed} to semantically represent a language description associated with the image data point in vector space. %
We use $C$-component GMMs to cluster our feature vectors---combined image features and paragraph vectors---and rank them. We consider the features which are closest to the center of cluster points to be the most informative data points and select them for training. 
 \renewcommand{\tabcolsep}{1pt}
\begin{table}[!ht]
\centering
\pgfplotstabletypeset[col sep=comma,
/color cells/max=1.0,
/color cells/min=0.0,
/color cells/textcolor=black,
columns/Color/.style={color cells},
columns/Shape/.style={color cells},
columns/Object/.style={color cells},
columns/Sampling/.style={string type,column type = {p{5cm}}},
/pgfplots/colormap={whitegray}{rgb255(0cm)=(255,255,255);rgb255(1cm)=(192,192,192)},
column type={>{\fontseries{bx}\selectfont\centering\arraybackslash}c},
every head row/.style={after row=\midrule},
every column/.style={
                column type/.add={}{|}
            },
]{
Sampling,Color,Shape,Object
\textit{Baseline: Random} , 0.75 , 0.19 , 0.49
\textit{Max-Log-Density-Based GMM Pool}, 0.82 , 0.25 , 0.62
\textit{{VL-GMM Sampling}}, 0.80 , 0.22 , 0.57
}
\caption{AUC summaries for each method's $F_1$ performance, grouped by the characteristic learned. Both  AL techniques performed better in characteristic grounding by selecting significant points from the pool.}
\label{tab:vlauc}
\end{table}

VL-GMM sampling (\cref{tab:vlauc}) outperformed a random baseline in learning groundings for color, shape, and object concepts, selecting the most diverse and informative data points at the earlier stages. VL-GMM consistently exhibited better performance; this makes intuitive sense, as this method uses language as well as image characteristics to select training data, and as such, has more information. While learning object types, VL-GMM selects only informative points at the initial stages, and initial performance is comparable to the baseline. After 50 data samples, it found diverse and representative data samples and ultimately outperformed all other sampling strategies. 

\subsection{Performance with Varying Data Size}
\begin{figure}[t]
\centering
    \includegraphics[width=.95\columnwidth]{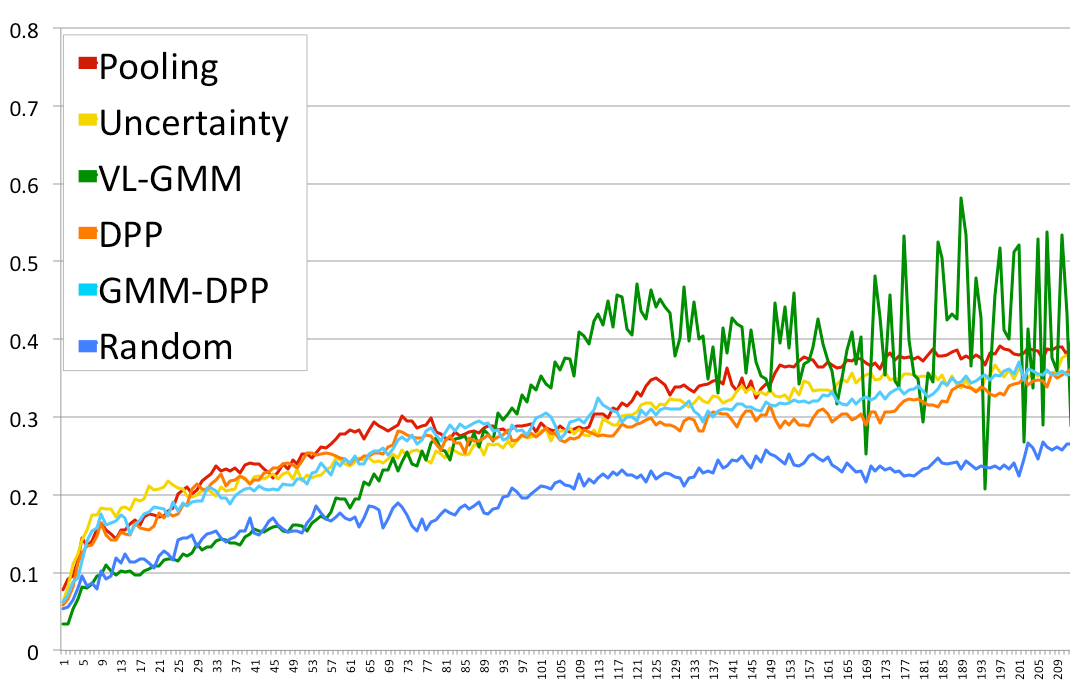}
	\caption{Performance of \textbf{visual classifiers} 
	for Object type
	as learning progresses with varying data size. Two hundred sixteen distinct object images and their annotations are used in training. $F_1$-score is shown on the \textit{y} axis, and number of data samples seen is shown on the $x$ axis.
	The VL-GMM approach shows promising performance in the more complex shape and objects classification problems. Still, the addition of noisy, highly varied descriptions in training affects the consistency in classification.
	}
    \label{fig:performance}
\end{figure}
In this experiment (\cref{fig:performance}), we try to mimic real-world human-robot learning that uses noisy, inconsistent, and limited data resources. For training, we used 216 distinct depth images and each image's description for training. We used the remaining 72 images for testing. The descriptions are highly noisy and varied. Most of them do not provide shape or object information. Our objective here is to understand how our active learning methods perform across varying amounts of available training data.  %
Due to space constraints, and to examine our methods under the ``harder'' setting where concepts are not frequently described, we show the results for ``object'' classification in \cref{fig:performance}. %
With highly varied and noisy features, all active learning algorithms could select diverse and important points from the pool using image features and perform better than the baseline for shape and object type words. However, linguistic variability within the description caused VL-GMM's performance to oscillate as it uses language while training. The results show that Max-Log-Density-Based pool sampling is consistently effective in all cases. This experiment also suggests that active learning algorithms that select informative and diverse points increase language acquisition quality especially when the training data is diverse and noisy.

\section{Analysis of Results}

The main conclusion we draw from our results is that the selection of the appropriate active learning method depends on the difficulty of the problem in terms of perceptual complexity, complexity and coverage of the language, and sparsity of objects in each class. However, we find that one or more active learning methods exist that can improve learning speed, overall performance, or both in all cases. We overall found GMMs to be a reliable choice for enhancing overall performance. These results are discussed further in this section.

\subsection{Method-Specific Findings}
GMM clustering with image features recovers a selection of data with both informative and diverse representation. This approach probabilistically clusters similar features in the same component. 
The uncertainty-based GMM is unable to effectively find patterns faster at initial stages in the dataset, when object classes are scattered in the visual space. Uncertainty sampling depends on the feature variability for finding uncertain points in GMM clustering, and the sampling selects the noisy outliers when the variability is greater. This finding echoes the performance reduction of uncertainty-based sampling in object feature spaces compared to pool-based approaches. 	

DPP variants of active learning methods with careful parameter tuning are well suited for selecting the most diverse points in the early stages of learning, which is appropriate when highly varied perceptual features make sample diversity important. Coverage of the more complex \Characteristic{shape} and \Characteristic{Object} attributes were attained significantly faster through these methods than random sampling. Visually varied datasets require more examples of concepts overall, in addition to requiring diversity. $k$-DPP sampling provides diverse samples from the dataset, and is proven sufficient for faster convergence of characteristic concepts. The DPP-based method is able to find the diverse data samples at the initial stages and provide faster convergence to the classification tasks with kernel descriptors as well as CNN features. 
However, \textit{representativeness}, along with \textit{diversity}, ensure consistent improvement. DPP sampling does not ensure representativeness and is not effective in the case of multi-colored, confusing samples. GMM-based structured DPPs provide breadth as well as diversity and perform well for simple and complex kernel descriptors data. However, this approach is weaker for CNN-based object classification, which may be because the process of selecting representative data adds unnecessary constraints. 

While the requirement for language in selecting data samples would be a limitation for large datasets, we found that sampling methods that could consistently augment the visual features with a small amount of language yielded improved grounded language systems. \newline

\noindent\textbf{Time Considerations.} For a dataset with N number of training data with D dimension, the determinantal point process computation requires $O(Nk + k ^2)$~\cite{kulesza2011k} if the eigen decomposition of the positive semi-definite kernel $K^{(0)}$ is available. And, eigen decomposition approximately takes $O(D^3)$.  Here $k$ denotes the size of subsets considered in DPP sampling. Similarly, the Gaussian Mixture model requires $O(D^3)$ to calculate weights that involve finding inverse and determinants. Since we calculate weights for every component and every data point, the overall time complexity of Max-Log-Density based approaches would be $O(C * N * D^3)$, where $N$ is the number of data points, and $C$ is the number of components. Structured DPP calculation involves GMM and DPP, so it requires $O(((N  + k) * k  + C ) * N * D^3)$ operations in total. After comparing the time complexities, Max-Log-Density based pool sampling seems suitable for large scale datasets.

\subsection{General Considerations} 
In all but the most trivial cases, random sampling from a dataset outperforms a sequential baseline. Since describing objects in order is a normal human behavior, this suggests that, lacking any other change, having an agent ask widely ranging questions in varying order may improve learning efficiency compared to passive learning. 

For cases in which neither visual percepts nor descriptive language varies widely, such as \Characteristic{color}, all active learning techniques are appropriate. We show that careful selection of \textit{informative} points is most critical under these circumstances: since the features are simple, the main consideration is to select representative data quickly, assuming that learning groundings (here, training visual classifiers) will proceed quickly. 

For visually differentiated, linguistically complex datasets, the importance of having a wide \textit{variety} of samples increases. 
DPPs~\cite{kulesza2012determinantal} are a class of `repulsive' processes designed for increasing diversity (see the discussion of $k$-DPP, above). Tuning with GMM parameters allows the DPP method to choose distinct, representative, and salient points in the data set in very early learning. Uncertainty-based max posterior GMM sampling performs well on complex data but does not perform as strongly for sparsely populated features.

We have shown that active learning techniques with carefully selected points reduce the amount of training data needed (see \cref{tab:auc}). We see that when dealing with more complex datasets, choosing diverse and meaningful points increases performance compared to choosing outliers. Our experiments have also shown that active learning helps set the right order of data points that can improve learning efficiency for both neural and non-neural visual features, and the addition of language features is not necessary for pool-based learning techniques to reduce the label cost. 

To summarize, GMM pool sampling, which decides certainty based on the density of the clustered data points, is the most reliable active learning choice for simple, complex, noisy, multi-colored, and highly varied datasets. It is consistently able to outperform random selection at least with 5\% increased predictive power. GMM uncertainty sampling is not a reliable choice in case of visual data with extremely noisy outliers. Logistic regression is the most robust classification model in modeling diverse limited data compared to SVM and MLP. DPP based and GMM based pool sampling produces good results in the case of neural and non-neural visual features. We observe that feature variability affects the selection techniques than the characteristics of the dataset. We believe that the vision-and-language sampling method considers the complexity and variance in visual features as well as the language features, and it aids in selecting the most diverse samples.

\section{Conclusion}
In this work, we present a thorough exploration of different active learning approaches to grounding unconstrained natural language in real-world sensor data. We demonstrate that active learning has the potential to reduce the amount of data necessary to ground language about objects, an active area of research in both NLP and robotics as well as machine learning from sparse data generally. We additionally provide suggestions for what approach may be suitable given the perceptual and linguistic complexity of a problem. Given our analysis of the causes of performance for different algorithms and cases, we believe these results will prove to generalize beyond the relatively simple data seen here, making it possible for these guidelines to apply to more complicated language grounding tasks in future.



\bibliography{references}
\bibliographystyle{IEEEtranS}

\end{document}